\documentclass[letterpaper, 10 pt, conference]{ieeeconf}  % Comment this line out if you need a4paper

\IEEEoverridecommandlockouts                              % This command is only needed if 
\usepackage{graphics}

\usepackage{amsfonts}
\usepackage{amsmath}

\usepackage{amssymb}
\usepackage[colorlinks, linkcolor=blue, citecolor=blue]{hyperref}
\usepackage{bm}

\usepackage{algorithm}
\usepackage{url}
\usepackage{algpseudocode}
\usepackage{epsfig}
\usepackage{graphicx}
\usepackage{caption}
\usepackage{subfigure}
\usepackage{mathrsfs}
\usepackage{array}
\usepackage{multirow}
\usepackage{multicol}
\usepackage{booktabs}
\usepackage{makecell}
\usepackage{xcolor}
\usepackage{amsfonts}
\usepackage{gensymb}
\usepackage[
    backend=biber,
    sorting=none
]{biblatex}
\usepackage{flushend}
\usepackage[flushleft]{threeparttable}
\usepackage{dsfont}
\newcommand*{\affmark}[1][*]{\textsuperscript{#1}}

\title{\LARGE \bf
Natural Humanoid Robot Locomotion with Generative Motion Prior
}

\author{Haodong Zhang\affmark[1,2]$^*$, Liang Zhang\affmark[1]$^*$, Zhenghan Chen\affmark[1,2], Lu Chen\affmark[1], Yue Wang\affmark[2], Rong Xiong\affmark[1,2]
\thanks{
$^*$These authors contributed equally to this work. $^1$Zhejiang Humanoid Robot Innovation Center. $^2$Zhejiang University. Rong Xiong is the corresponding author. {\tt\small rxiong@zju.edu.cn}.
This work was supported by the ''Jian Bing Ling Yan+`` Technology Plan 2024C01SA202315.
}
}

\begin{document}

\maketitle
\thispagestyle{empty}
\pagestyle{empty}
% \def\thefootnote{*}\footnotetext{These authors contributed equally to this work}
% \def\thefootnote{1}\footnotetext{Zhejiang Humanoid Innovation Center}\def\thefootnote{2}\footnotetext{Zhejiang University}
% \def\thefootnote{\dag}\footnotetext{Corresponding author: rxiong@zju.edu.cn}

%%%%%%%%%%%%%%%%%%%%%%%%%%%%%%%%%%%%%%%%%%%%%%%%%%%%%%%%%%%%%%%%%%%%%%%%%%%%%%%%
\begin{abstract}

Natural and lifelike locomotion remains a fundamental challenge for humanoid robots to interact with human society. However, previous methods either neglect motion naturalness or rely on unstable and ambiguous style rewards. In this paper, we propose a novel Generative Motion Prior (GMP) that provides fine-grained motion-level supervision for the task of natural humanoid robot locomotion. To leverage natural human motions, we first employ whole-body motion retargeting to effectively transfer them to the robot. Subsequently, we train a generative model offline to predict future natural reference motions for the robot based on a conditional variational auto-encoder. During policy training, the generative motion prior serves as a frozen online motion generator, delivering precise and comprehensive supervision at the trajectory level, including joint angles and keypoint positions. The generative motion prior significantly enhances training stability and improves interpretability by offering detailed and dense guidance throughout the learning process. Experimental results in both simulation and real-world environments demonstrate that our method achieves superior motion naturalness compared to existing approaches. Project page can be found at \url{ https://sites.google.com/view/humanoid-gmp}
\end{abstract}

%%%%%%%%%%%%%%%%%%%%%%%%%%%%%%%%%%%%%%%%%%%%%%%%%%%%%%%%%%%%%%%%%%%%%%%%%%%%%%%%
\section{INTRODUCTION}

Humanoid robots are a type of universal embodied agent capable of performing various manipulation and locomotion tasks, including carrying boxes \cite{zhang2024wococo} and traversing complex terrains \cite{zhuang2024humanoid}.
Due to their human-like morphology, humanoid robots are inherently more suitable for interacting with humans.
Researchers have empowered humanoid robots to shake hands with humans \cite{cheng2024expressive} and perform expressive movements \cite{he2025asap}.
Locomotion is one of the most fundamental abilities of humanoid robots, and endowing them with natural human-like locomotion capabilities is crucial for helping them interact with human society and enter households.
% better

Existing methods have been proposed to equip humanoid robots with fundamental locomotion abilities.
Model Predictive Control (MPC) methods utilize precise mathematical models and optimize control variables to achieve whole-body balance control of robots \cite{ishihara2019full}.
Reinforcement Learning (RL) methods develop control policies by maximizing rewards when interacting with the simulation environment, and subsequently transfer these policies to physical robotic systems \cite{zhuang2024humanoid,wei2023learning}.
However, most of these approaches overlook the human-likeness of locomotion, often generating unnatural robotic movements characterized by bent-leg gaits, mechanical motion patterns, and a general lack of human-like fluidity. Such limitations hinder the seamless integration of humanoid robots into human-centric environments and their effective interaction with human society.

\begin{figure}[t]
\centering
% \vspace{6pt}
\includegraphics[width=0.9\linewidth]{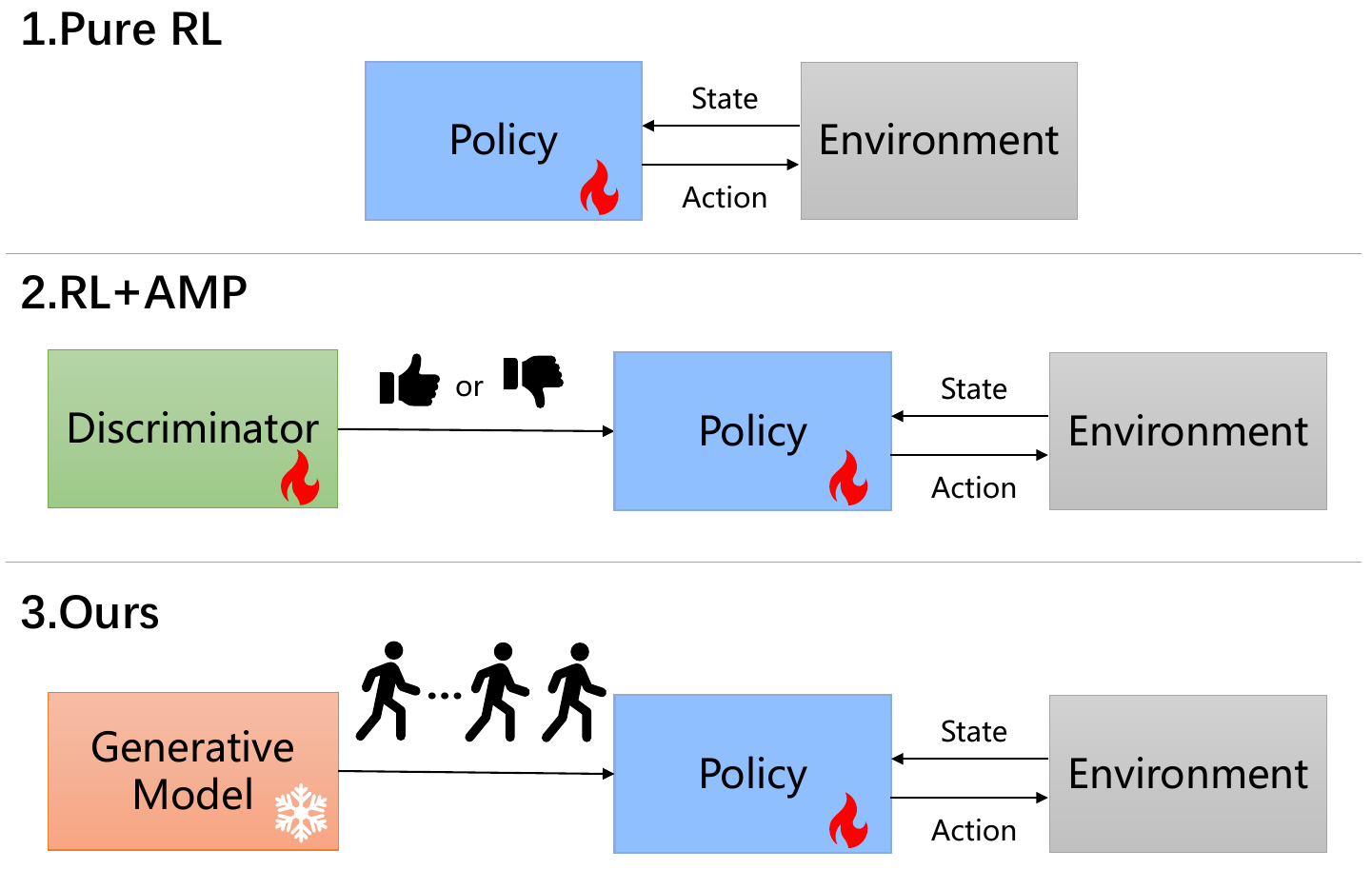}
\caption{Illustration of different methods. (a) Pure RL only considers task objectives and neglects motion naturalness. (b) RL+AMP simultaneously trains the RL policy with the discriminator to provide an ambiguous scalar style reward and suffers from training instability.
% that indicates motion quality.
(c) Our method incorporates a frozen generative model to predict natural whole-body reference motion trajectories for the robot and provides more stable and detailed motion guidance.}
\label{fig:introduction}
\vspace{-16pt}
\end{figure}

To address the task of human-like natural locomotion, the key insight is to learn from human motion data. Human motion data contains diverse information of natural transitions and motion patterns. Utilizing human motion data is an efficient way to learn natural humanoid robot locomotion.
Recent advancements have introduced Adversarial Motion Prior (AMP), which is derived from Generative Adversarial Networks (GANs) \cite{goodfellow2014generative} and applied to generate lifelike motions for physically simulated characters \cite{peng2021amp}.
This approach simultaneously learns a policy network with a discriminator network to evaluate the style of movements and provide style rewards for the policy network.
For humanoid robot locomotion, researchers have introduced AMP to indicate whether the robot motions have similar locomotion styles to the reference demonstrations \cite{zhang2024whole,tang2024humanmimic}. The discriminator treats the retargeted human motion data as real samples and the robot motions of the RL control policy as negative samples.
% The goal of the control policy is to confuse the discriminator network. 
% Tang et al. proposed to improve AMP's training stability with a soft boundary constraint and achieved natural locomotion in the simulation environment \cite{tang2024humanmimic}.
Although these methods can implicitly imitate human motion styles through AMP, the adversarial training notoriously suffers from inherent instability and mode collapse \cite{goodfellow2014generative,arjovsky2017wasserstein,arjovsky2017towards}. The co-training of the RL control policy and discriminator easily gets stuck when the discriminator does not learn well and fails to provide effective guidance. Furthermore, the style reward computed by the discriminator is an ambiguous scalar value, which lacks granularity and interpretability, further limiting its ability to provide detailed guidance.

% While learning the control policy, a discriminator network is also learned. 
% However, this approach requires concurrent training of both the control policy and the discriminator network, leading to potential issues of training instability and mode collapse.

To efficiently leverage human motion data to guide humanoid robots in learning natural locomotion, we present a novel learning-based method named Generative Motion Prior (GMP), which utilizes a generative model to online synthesize human-like reference motion trajectories for the robot and provide granular guidance signals for the natural locomotion behaviors as Fig.~\ref{fig:introduction}.
Specifically, we first perform whole-body motion retargeting to transform human natural motion data to the humanoid robot as reference demonstrations. Subsequently, we train a motion generation network based on a Conditional Variational Auto-encoder (CVAE), which is capable of predicting future natural robot motions in an auto-regressive manner.
The motion generator synthesizes the subsequent robot pose based on the current robot pose and the user velocity command.
To encourage human-like robot movements, we introduce several motion guidance rewards to provide fine-grained and dense guidance at the motion level, including joint angles and keypoint positions. To facilitate training efficiency and stability, the motion generator is trained offline and kept frozen during RL policy training. To the best of our knowledge, we are the first to utilize the capabilities of a motion generator for the task of learning natural humanoid robot locomotion.

% By pretraining a conditional variational autoencoder, we generate future human-like walking trajectories to provide better guidance for reinforcement learning.
% Specifically, we  on human walking data and learn a generative model from the retargeted data. This model can predict the next frame of motion  based on the robot's previous state, including Com velocity and joint angles, encouraging the robot's movements to be more human-like through dense guided rewards.

Our contributions can be summarized as follows:

\begin{itemize}

    \item introduce an innovative framework that incorporates the generative motion prior to act as a frozen expert model and provides robust and stable guidance for the task of learning human-like humanoid robot locomotion.
    
    % which is composed of reinforcement learning and generative motion prior to act as direct supervision that is more interpretable. .

    \item propose to predict fine-grained whole-body reference motion trajectories for robots using a generative model and design motion guidance rewards to guide the RL control policy in a granular and interpretable manner.
    % help the learning process of

    \item conduct experiments on the task of natural humanoid locomotion in both physics simulation and real-world environments and achieves the state-of-the-art performance in motion naturalness.
    % , convergence speed and training stability.

\end{itemize}

% stability
% detail whole-body

\section{RELATED WORKS}

\subsection{Learning Based Humanoid Whole-body Control}

Reinforcement learning has demonstrated remarkable robustness and efficiency in quadruped robot control \cite{rudin2022learning,kumar2021rma}, and its application has recently been extended to whole-body control of humanoid robots. Several notable advancements have been made in this domain.
Jeon et al. \cite{jeon2023benchmarking} introduced potential-based reward shaping to enhance the learning process and accelerate convergence.
Wei et al. \cite{wei2023learning} developed a learning-based humanoid locomotion controller incorporating gait-conditioned Rapid Motor Adaptation for improved disturbance resistance and terrain adaptability.
Van et al. \cite{van2024revisiting} proposed a standing and walking controller capable of command following and disturbance recovery with energy efficiency.
He et al. \cite{he2024hover} proposed a unified control policy supporting multiple control modes, including kinematic position tracking, local joint angle tracking, and root tracking.
Radosavovic et al. \cite{radosavovic2024humanoid} formulated humanoid control as a next-token prediction problem, employing a causal transformer to learn from sensorimotor trajectories.
Zhang et al. \cite{zhang2024wococo} presented a whole-body control framework that decomposes tasks into multiple contact stages.
Zhuang et al. \cite{zhuang2024humanoid} implemented a vision-based whole-body control policy enabling humanoid robots to navigate challenging obstacles.
However, while these methods have advanced various manipulation and locomotion tasks, they predominantly overlook the motion naturalness, often resulting in robotic movements that appear ungraceful and lack human-like fluidity.

% However, these methods focus on different manipulation or locomotion tasks, and neglect the motion naturalness of humanoid robots, resulting in ungraceful and awkward robot movements.

\subsection{Imitation from Human Motion}

Mimicking human data has emerged as an effective paradigm for enabling humanoid robots to acquire complex skills.
He et al. \cite{he2024learning} developed a reinforcement learning framework for real-time humanoid robot teleoperation using RGB cameras.
Fu et al. \cite{fu2024humanplus} introduced a system for humanoid robots to learn both motion and autonomous skills from human data.
Cheng et al. \cite{cheng2024expressive} proposed decoupling whole-body motion into upper-body imitation and lower-body velocity tracking.
Lu et al. \cite{lu2024mobile} implemented a predictive motion prior for precise upper-body control and robust lower-body locomotion.
Ji et al. \cite{ji2024exbody2} created a whole-body motion tracking framework capable of generalizing across diverse motion inputs.
Jiang et al. \cite{jiang2024harmon} proposed to generate whole-body motions from natural language descriptions for humanoid robots and refine the motions with vision-language models.
While these works focus on learning complex human skills or imitating expressive human motions, our study specifically addresses the challenge of achieving natural locomotion in humanoid robots.
For the task of natural bipedal locomotion learning, previous researchers have explored adversarial motion prior techniques \cite{zhang2024whole,tang2024humanmimic}. 
Zhang et al. \cite{zhang2024whole} developed a whole-body humanoid locomotion system using adversarial motion prior, where a discriminator assesses the similarity between human reference motions and robot motions. Tang et al. \cite{tang2024humanmimic} enhanced this approach by incorporating an adversarial Wasserstein critic with soft boundary constraints to stabilize training and prevent model collapse.
However, these adversarial training methods are inherently unstable, and the resulting style rewards often lack detailed guidance and interpretability.
In contrast, our proposed method provides more granular and sufficient guidance throughout the learning process, addressing these limitations while maintaining motion naturalness.

\begin{figure*}[t]
\centering
\vspace{6pt}
\includegraphics[width=0.9\linewidth]{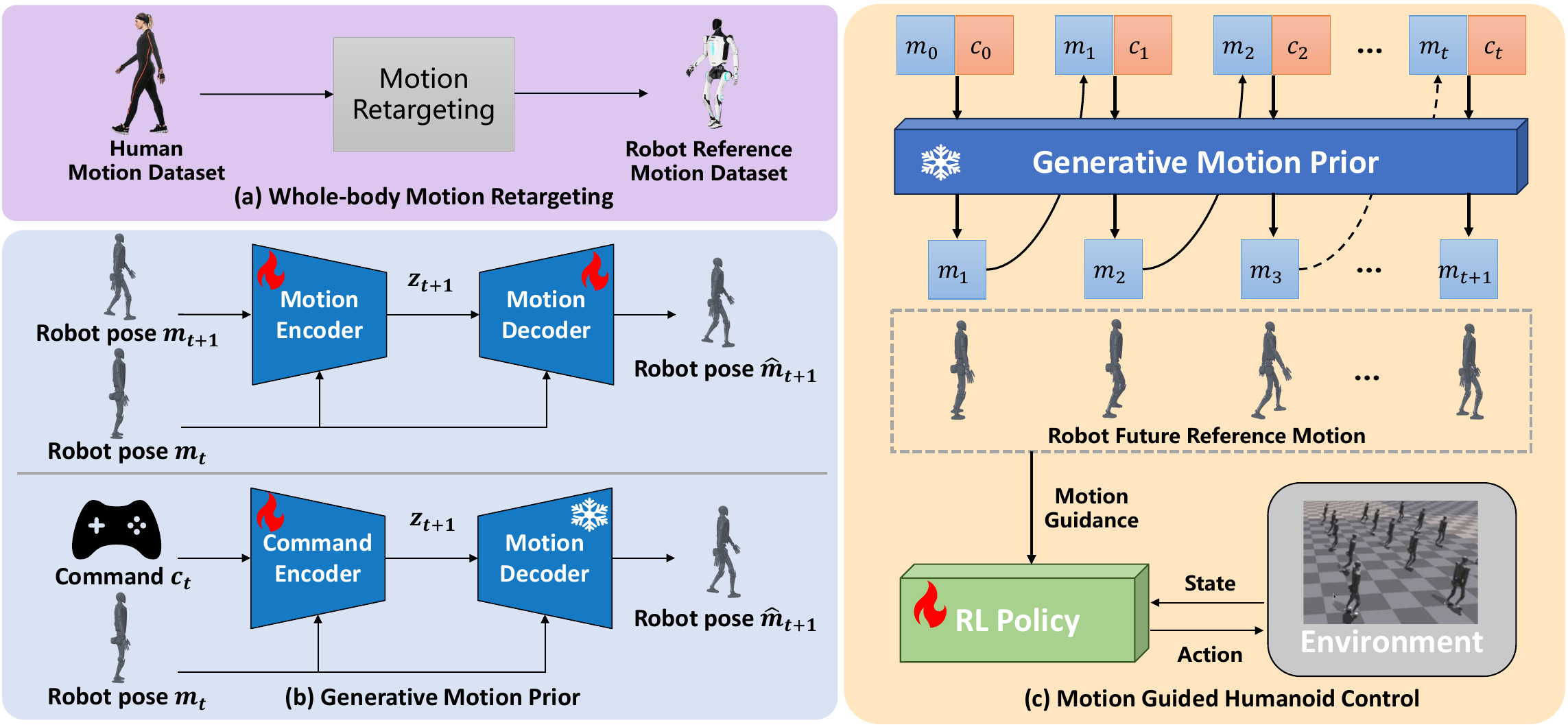}
\caption{Overall framework. (a) First, we transform the human motion dataset to the robot reference motion dataset by whole-body motion retargeting. (b) Utilizing the retargeted robot motions, we train a generative motion prior that can predict natural robot pose $\boldsymbol{m}_{t+1}$ at the next frame based on current robot pose $\boldsymbol{m}_{t}$ and the user velocity command $\boldsymbol{c}_{t}$. (c) The frozen generative motion prior performs online motion generation in an auto-regressive manner to synthesize the robot future reference motion and guide the RL policy to learn natural locomotion.}
\label{framework}
\vspace{-15pt}
\end{figure*}

\subsection{Human Motion Generation}

Shi et al. \cite{shi2024interactive} proposed an auto-regressive motion diffusion model and achieved real-time interactive character control via hierarchical reinforcement learning.
Tevet et al. \cite{tevet2023human} introduced a classifier-free diffusion-based generative model for the human motion synthesis tasks, including text-to-motion, action-to-motion, and unconditioned generation.
Guo et al. \cite{guo2024momask} combined the residual vector quantization and generative masked transformers for high-quality text-driven 3D human motion generation.
In comparison, this paper introduces a natural robot motion generator to guide the learning process, rather than generating human motion.

\section{METHOD}

\subsection{Overview}

In this section, we present a natural humanoid robot locomotion algorithm guided by generative motion prior, as illustrated in Fig.~\ref{framework}. The framework begins by mapping naturally coordinated human motion data to the humanoid robot through whole-body motion retargeting. Based on the retargeted robot reference motion data, we construct a generative motion prior model that is capable of generating the next frame motion according to desired commands. During the training process of the RL control policy, the frozen generative model synthesizes future whole-body reference motion trajectories for the robot in real time, providing motion guidance to the RL policy network and encouraging the humanoid robot to achieve natural human-like locomotion.

% captures the probability distribution from the current frame motion $\boldsymbol{m}_t$ to the next frame motion $\boldsymbol{m}_{t+1}$, and

\subsection{Whole-body Motion Retargeting}

The objective of whole-body motion retargeting is to transform human motion data into humanoid robots. In this section, we extend the motion retargeting method introduced in \cite{zhang2022kinematic} from the dual-arm robot to the humanoid robot. To keep motion similarity and satisfy kinematic robot constraints, whole-body motion retargeting is formulated as an optimization problem. The objective functions contain three terms as Eq.~\ref{obj}, including vector similarity loss, foot contact loss, and motion smoothness loss, where $\alpha$, $\beta$, and $\gamma$ are respective weights.
% , and root velocity loss
\begin{equation}
\label{obj}
    L = \alpha L_{vec} + \beta L_{foot} + \gamma L_{smo}% + L_{root}
\end{equation}

The vector similarity loss $L_{vec}$ encourages the directional vectors formed by key joints to maintain consistent between the source human motion and the target robot motion, including from shoulders to elbows, from elbows to wrists, from hips to knees, and from knees to ankles. Let $V$ be the vector set, then it is defined as:

\begin{equation}
    L_{vec} = \sum_{j \in V}\left|\left|\boldsymbol{v}_j^{human}-\boldsymbol{v}_j^{robot}\right|\right|^2_2
\end{equation}

The foot contact loss $L_{foot}$ encourages the humanoid robot to maintain foot contact relationship with the ground the same as the source human motion. The foot contact is represented as binary signals by detecting whether the foot height and velocity are close to zero. Let $C$ be the foot set in contact with the ground, $h_j$ be the height and $\dot{\boldsymbol{p}}_j$ be the velocity of foot $j$, then it is given as:

\begin{equation}
    L_{foot}=\sum_{j \in C}\left|\left|h_j\right|\right|^2_2+\left|\left|\dot{\boldsymbol{p}}_j\right|\right|^2_2
\end{equation}

The motion smoothness loss $L_{smo}$ encourages smooth robot motions and penalizes the acceleration of joint angles. Let $\dot{\boldsymbol{q}}_{t+1}$ be the joint velocity at frame $t+1$ and $\dot{\boldsymbol{q}}_{t}$ be the joint velocity at frame $t$, then it is defined as:

\begin{equation}
    L_{smo} = \left|\left|\dot{\boldsymbol{q}}_{t+1} - \dot{\boldsymbol{q}}_{t}\right|\right|^2_2
\end{equation}

% The base velocity loss $L_{base}$ encourages the base velocity to be similar to the source human motion, normalized by the leg length. Let $v_{base}^{human}$ and $l^{human}$ be the base velocity and leg length of the human and $v_{base}^{robot}$ and $l^{robot}$ be the corresponding variables of the robot, the it is given as:

% \begin{equation}
%     L_{base} = \left|\left|\frac{v_{base}^{human}}{l^{human}}-\frac{v_{base}^{robot}}{l^{robot}}\right|\right|^2_2
% \end{equation}

\subsection{Generative Motion Prior}

\begin{figure}[t]
\centering
\vspace{6pt}
\includegraphics[width=0.8\linewidth]{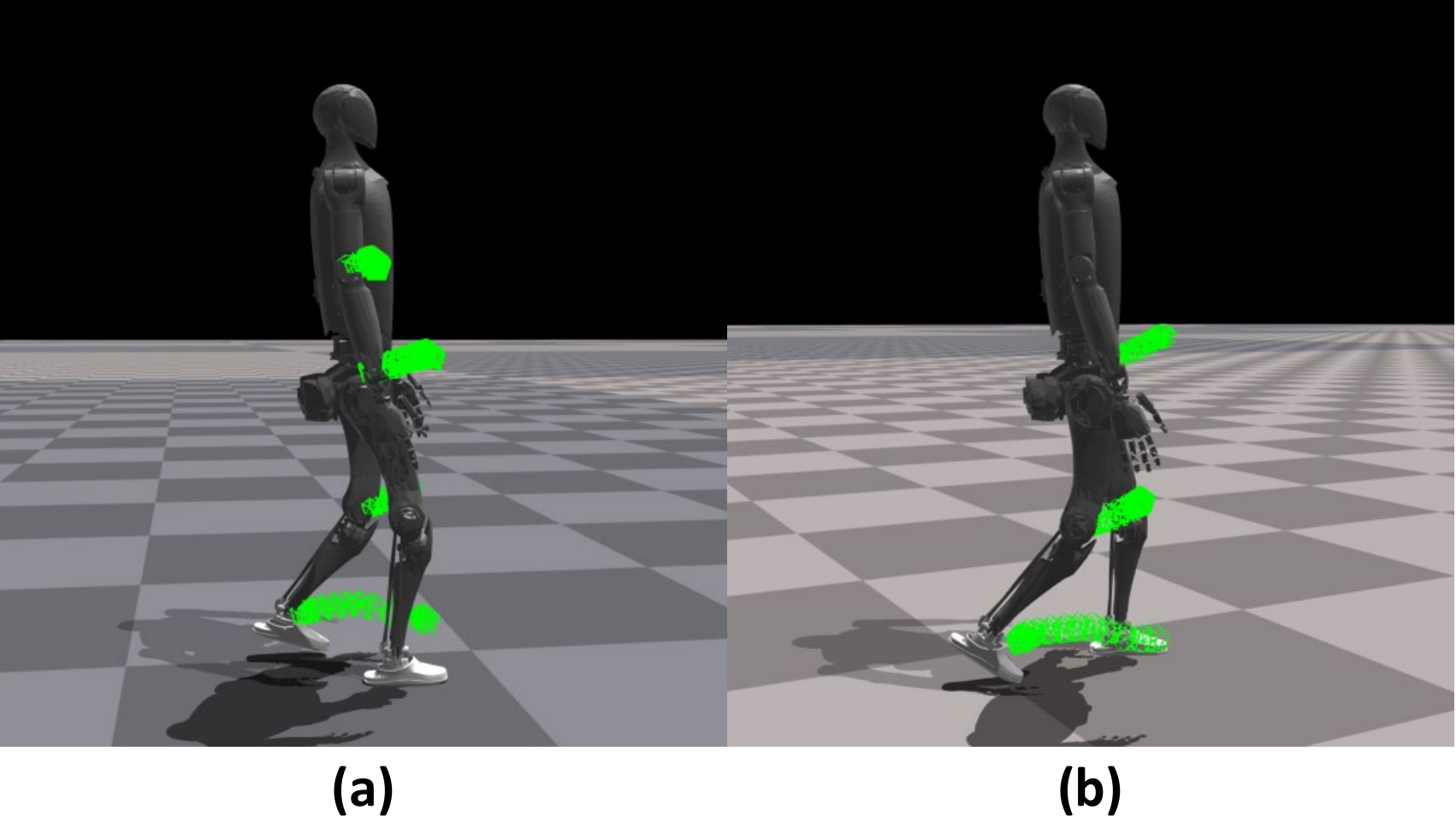}
\caption{Visualization of future robot reference motion predicted by the generative motion prior, with green dots indicating the keypoint positions for the subsequent twelve frames. (a) Trajectories of the left knee and ankle alongside the right elbow and wrist. (b) Trajectories of the right knee and ankle together with the left elbow and wrist.}
\label{fig:guidance-visualization}
\vspace{-12pt}
\end{figure}

Given the natural robot motions retargeted from human data, we introduce a generative motion prior, which learns to predict future human-like robot motion trajectories to guide the humanoid robot locomotion. The generated robot reference motion provides more granular and interpretable whole-body reference information, including the robot joint angles, keypoint positions, base velocities, and base height.
Furthermore, the generative prior model is trained offline, therefore it is capable of serving as a frozen expert model and providing more stable guidance during RL control policy training, which avoids the training instability problem of adversarial training \cite{goodfellow2014generative,arjovsky2017wasserstein,arjovsky2017towards}. As shown in Fig.~\ref{fig:guidance-visualization}, the generative motion prior is able to predict natural reference motion trajectories for the humanoid robot.

\noindent\textbf{Conditional Variational Auto-encoder.}
To generate future human-like reference motions for the humanoid robot, we introduce a conditional variational auto-encoder (CVAE) framework to model the probabilistic distribution of the next-frame robot reference pose $\boldsymbol{m}_{t+1}$ conditioned on the current-frame reference pose $\boldsymbol{m}_{t}$, denoted as $p(\boldsymbol{m}_{t+1}|\boldsymbol{m}_{t})$. Specifically, the robot reference pose at frame $t$ is represented as $\boldsymbol{m}_t = (\boldsymbol{v}^{base}_t, \boldsymbol{w}^{base}_t, \boldsymbol{q}_t, \boldsymbol{p}^{key}_t, \boldsymbol{v}^{key}_t)$, which encapsulates the base linear velocity $\boldsymbol{v}^{base}_t$, base angular velocity $\boldsymbol{w}^{base}_t$, joint angles $\boldsymbol{q}_t$, keypoint positions $\boldsymbol{p}^{key}_t$, and keypoint velocities $\boldsymbol{v}^{key}_t$. This comprehensive representation captures both the global and local motion characteristics, enabling the generation of natural and coherent reference motions for humanoid robots.

The conditional variational auto-encoder consists of two core components: a motion encoder and a motion decoder, as illustrated in Fig.~\ref{framework}. The motion encoder $f_\theta$, parameterized by $\theta$, establishes a probabilistic mapping $p(\boldsymbol{z}_{t+1}|\boldsymbol{m}_{t+1},\boldsymbol{m}_{t})$. As depicted in Eq.~\ref{eq:encoder}, the motion encoder processes the subsequent frame's robot pose $\boldsymbol{m}_{t+1}$ as its primary input while conditioning on the current robot pose $\boldsymbol{m}_{t}$, ultimately projecting this information into a latent space characterized by a multivariate Gaussian distribution.

\begin{equation}
\label{eq:encoder}
    \boldsymbol{z}_{t+1} = f_\theta(\boldsymbol{m}_{t+1},\boldsymbol{m}_{t})
\end{equation}

On the other hand, the motion decoder $f_\phi$, parameterized by $\phi$, establishes the inverse transformation $p(\boldsymbol{m}_{t+1}|\boldsymbol{z}_{t+1},\boldsymbol{m}_{t})$. As depicted in Eq.~\ref{eq:decoder}, the motion decoder reconstructs the predicted next-frame robot pose $\hat{\boldsymbol{m}}_{t+1}$ by synthesizing information from the latent vector $\boldsymbol{z}_{t+1}$ conditioning on the current robot pose $\boldsymbol{m}_{t}$.

\begin{equation}
\label{eq:decoder}
    \hat{\boldsymbol{m}}_{t+1} = f_\phi(\boldsymbol{z}_{t+1},\boldsymbol{m}_{t})
\end{equation}

% TODO: beta VAE
The training objective of the conditional variational auto-encoder consists of the motion reconstruction loss and the KL divergence loss. 
The motion reconstruction loss is designed to minimize the discrepancy between the reconstructed motion and the original input. This loss is computed using the mean squared error (MSE), as detailed in Eq.~\ref{eq:reconstruction}. 
On the other hand, the KL divergence loss regularizes the latent space by encouraging the latent vectors to conform to a multivariate Gaussian distribution.
To enhance the model's capability in generating accurate long-sequence robot motions, we employ scheduled sampling \cite{bengio2015scheduled} during the training process. It involves randomly substituting the ground truth data with the model's predicted robot pose as the input for the next frame prediction.
% This regularization promotes a structured and interpretable latent representation, which is essential for generating diverse and coherent motion sequences.
% The model is trained to auto-regressively generate future eight frames of robot reference motion.
% By gradually increasing the reliance on the model's own predictions, scheduled sampling mitigates the exposure bias problem, where the model becomes overly dependent on ground truth data during training but must rely on its own predictions during inference. This approach significantly improves the robustness and accuracy of long-sequence motion generation.

% By combining these loss functions and the scheduled sampling strategy, the model is optimized to produce precise and natural motion sequences while maintaining a well-structured latent space.

% The generative prior is trained with retargeted robot motions. It generates reference motions for the next eight frames based on the current state.   The training loss is shown in Eq.~\ref{autoencoder}. We predict future eight frames robot poses. 

\begin{equation}
\label{eq:reconstruction}
    L_{rec} =\left|\left|\hat{\boldsymbol{m}}_{t+1}-\boldsymbol{m}_{t+1}\right|\right|_2^2
    %+KL(f_\theta(\boldsymbol{z}_t|\boldsymbol{m}_t,s_{t-1})|\mathcal{N}(0,\,1)|)
\end{equation}

\noindent\textbf{Auto-Regressive Motion Generation.} After training, the latent space of the conditional variational auto-encoder can be effectively employed to sample the robot's motion for the subsequent frame.
Specifically, the latent vector $\boldsymbol{z}_{t+1}$ is sampled from a multivariate Gaussian distribution, and the motion decoder generates the predicted next frame robot pose $\hat{\boldsymbol{m}}_{t+1}$, conditioned on the current robot pose $\boldsymbol{m}_{t}$, as Eq.~\ref{eq:decoder}.
In an auto-regressive manner, we can leverage the motion decoder to predict natural robot reference motions of arbitrary length. 
For instance, the decoder utilizes the predicted robot pose $\hat{\boldsymbol{m}}_{t+1}$ as input to forecast the subsequent pose $\hat{\boldsymbol{m}}_{t+2}$, as Eq.~\ref{eq:autoregressive}. By iteratively applying this procedure $N$ times, a long-horizon sequence of natural robot motions can be synthesized, represented as \( \{\hat{\boldsymbol{m}}_{t+1}, \hat{\boldsymbol{m}}_{t+2}, \ldots, \hat{\boldsymbol{m}}_{t+N}\} \).
This approach ensures the generation of natural and realistic robot reference motion trajectories.
% TODO: KINEMATIC
% We discard the motion encoder and  .
% the motion decoder $f_\phi$ has the capability to 

% a latent vector $\boldsymbol{z}_{t+1}$ is sampled from the Gaussian distribution with a mean of 0 and a variance $\sigma$, as Eq.~\ref{eq:sample}. Then

% In an auto-regressive manner, the motion decoder $f_\phi$ is capable of predicting arbitrary length of robot reference motion. For example, we can use the predicted robot pose $\hat{\boldsymbol{m}}_{t+1}$ as input to predict the subsequent robot pose $\hat{\boldsymbol{m}}_{t+2}$ as Eq.~\ref{eq:autoregressive}. The latent vector $\boldsymbol{z}_{t+1}$ is sampled from the Gaussian distribution. 
% We could follow this process for $N$ times to synthesize long-horizon natural robot motion sequence $\{\boldsymbol{m}_t, \boldsymbol{m}_{t+1}, \boldsymbol{m}_{t+2}, ...,\boldsymbol{m}_{t+N}\}$.
% This process could be done for multiple times. 
% Therefore, we are able to generate 

\begin{gather}
    % \label{eq:sample}
    % \boldsymbol{z}_{t+1} \sim \mathcal{N}(0,\,\sigma^{2})\\
    \label{eq:autoregressive}
    \hat{\boldsymbol{m}}_{t+2} = f_\phi(\boldsymbol{z}_{t+2},\hat{\boldsymbol{m}}_{t+1})
\end{gather}

%  and the blue dots denote current keypoint positions

\noindent\textbf{Command Encoder.}
In the auto-regressive process of robot motion generation, the prediction of future motion trajectories is generated by randomly sampling from the latent space, resulting in diverse but uncontrollable velocities of robot reference motion.
For example, from an initial motion frame, the generated future motion trajectory may speed up or slow down, leading to undesired robot velocities that deviate from intended commands.
To enable precise velocity control in response to user commands, we introduce a command encoder $f_\psi$, where $\psi$ represents the learnable parameters.
This command encoder is specifically designed to generate appropriate latent vectors that ensure the generated motion adheres to the desired velocity command.
It takes the velocity command $\boldsymbol{c}_t$ and current robot pose $\boldsymbol{m}_{t}$ as input and subsequently outputs a latent vector $\boldsymbol{z}_{t+1}$ as Eq.~\ref{eq:command}, which is then used as the input of the motion decoder.
We follow the training pipeline introduced in \cite{shi2024interactive,won2022physics,ling2020character}, which is formulated as learning a hierarchical policy that takes the latent vector as the action space.
During training, we keep the parameters of the motion decoder frozen and optimize the parameters of the command encoder as Fig.~\ref{framework}. 
This approach facilitates us to utilize the command encoder to guide the generation of the future robot reference motion in response to the specified velocity command.
% conditioned on historical motion data.
% Meanwhile, the latent vectors are randomly sampled from a Gaussian distribution, 
% resulting in uncontrollable robot velocity.
% that guides the generation of the next frame's robot pose in accordance with the specified velocity command as Eq.~\ref{command}.
% The latent vector is then used as the input of the motion decoder.
% To generate guidance motion according to velocity commands, we formulate the velocity control task as learning a control policy which takes the latent vector as an action space as \cite{shi2024interactive,won2022physics,ling2020character}.

% is an autoregressive model that generates future states based on historical states. However, the velocity generated by randomly sampling noise from a Gaussian distribution are uncontrollable. To provide more effective guidance, it is necessary to generate corresponding actions according to speed commands. 

\begin{equation}
\label{eq:command}
    \boldsymbol{z}_{t+1} = f_\psi(\boldsymbol{c}_t,\boldsymbol{m}_{t})
\end{equation}

% The module generates a future full-body motion trajectory based on past historical states and the current speed command as input.
% Add an additional network module that takes the speed command as input and outputs latent variables, replacing random sampling. 
% We utilize a three-layer MLP parameterized as $f_\psi$ to 

% Finally, we takes the velocity command and last robot state as input and predicts the next frame robot pose

% \begin{equation}
%     \boldsymbol{m}_{t+1} = f_\phi((f_\psi(\boldsymbol{c}_t,\boldsymbol{m}_{t}), \boldsymbol{m}_{t})
% \end{equation}

\subsection{Motion Guided Humanoid Control}

We formulate the humanoid robot control task as a goal-conditioned reinforcement learning problem, where the objective is to find the optimal control policy $\pi^*$ that simultaneously tracks the user velocity commands and maintains human-like motion naturalness.
The state $\boldsymbol{s}_t$ is composed of three components: (1) the robot's proprioceptive information $\boldsymbol{s}^p_t$, (2) the user velocity command $\boldsymbol{c}_t$, and (3) the reference motion $\boldsymbol{m}_{t+1}$ predicted by the frozen generative motion prior.
The motion prior acts as an online motion generator, providing crucial motion guidance signals to the RL control policy.
The proprioceptive state $\boldsymbol{s}^p_t$ encompasses joint positions $\boldsymbol{q}_t$, joint velocities $\dot{\boldsymbol{q}}_t$, base angular velocity $\boldsymbol{\omega}_t$, the last action $\boldsymbol{a}_{t-1}$, and the projected gravity vector $\boldsymbol{g}$.
The RL policy processes the state $s_t$ and computes target joint positions as outputs. These positions are subsequently converted into torque commands through the PD controller.
For policy optimization, we employ Proximal Policy Optimization (PPO) \cite{schulman2017proximal} to maximize the expected cumulative reward $\mathbb{E}\left[\sum_{t=1}^T\gamma^{t-1}r_t\right]$, where $\gamma$ is the discount factor.

% To promote the motion naturalness and accomplish the task objective that tracks the velocity command, the reward function contains three terms as Eq.~\ref{eq:reward}, including the task-agnostic motion guidance reward $r_{guidance}$, the task reward $r_{task}$, and the regularization reward $r_{reg}$.

To enhance the naturalness of the robot's motion while simultaneously achieving the task objective of velocity command tracking, the reward function is composed of three components, as Eq.~\ref{eq:reward}. These include: (1) the task-agnostic motion guidance reward $r_{guidance}$, which ensures the robot's movements align with natural, human-like motion patterns; (2) the task reward $r_{task}$, which directly incentivizes the fulfillment of the velocity tracking objective; and (3) the regularization reward $r_{reg}$, which imposes constraints to prevent undesirable behaviors and ensure stability. Together, these terms form a comprehensive reward structure that balances motion quality and task performance.

\begin{equation}
\label{eq:reward}
    r = r_{guidance} + r_{task} + r_{reg}
\end{equation}

% This comprehensive framework integrates advanced reinforcement learning techniques with precise physical modeling, enabling the development of robust and natural humanoid robot control systems. The inclusion of both proprioceptive feedback and generative motion guidance ensures that the resulting motions are both physically feasible and perceptually natural, addressing key challenges in humanoid robotics.

% The humanoid robot control task is formulated as a Markov Decision Process $\mathcal{M}=(\mathcal{S},\mathcal{A},\mathcal{T},\mathcal{R},\gamma,)$, consisting of state space $s_t\in\mathcal{S}$, action space $a_t\in\mathcal{A}$, transition probability $\mathcal{T}$, reward function $r_t\in\mathcal{R}$.
% The objective of the control policy is .
% ${\pi}(a_t|s_t,m_t)$

% The goal of the control policy is to track the user velocity commands as well as maintain human-like motion.
 
% we also concatenate the reference keypoint positions, reference keypoint velocities, reference joint angles.
% seems like motion tracking 

\noindent\textbf{Motion Guidance Reward.}
The generative motion prior serves as a predictive model capable of forecasting future whole-body reference motions for the humanoid robot, including both joint positions and keypoint positions. This reference motion provides fine-grained guidance signals essential for generating natural and plausible locomotion behaviors. To effectively leverage this prior, we design motion guidance rewards that quantify the discrepancies between the robot's actual states and the reference trajectory. Specifically, we introduce two distinct reward components: the joint angle guidance reward and the keypoint position guidance reward, as Eq.~\ref{eq:guidance}, to promote human-like motion characteristics in the robot's behavior.

 % (joint positions, velocities, centroid height, speeds, etc.). 

\begin{equation}
\label{eq:guidance}
    r_{guidance} = r_{dof} + r_{keypos}% + r_{height}
\end{equation}

The joint angle guidance reward encourages the robot's joint angles to align closely with those of the reference motion. Let $\boldsymbol{q}$ and $\boldsymbol{q}_{ref}$ denote the joint angles of the robot and the reference motion, respectively. It is defined as:

\begin{equation}
    r_{dof} = e^{-0.7\left|\left|{\boldsymbol{q}-\boldsymbol{q}_{ref}}\right|\right|}
\end{equation}

The keypoint position guidance reward encourages the keypoint positions, such as end-effectors (e.g., wrists and ankles), of the robot to closely match those of the reference motion. The keypoint positions are transformed into the robot's local coordinate system relative to the robot base. Let $\boldsymbol{p}$ and $\boldsymbol{p}_{ref}$ represent the keypoint position of the robot and reference motion, respectively. It is given as:

\begin{equation}
    r_{keypos} = e^{-0.7\left|\left|{\boldsymbol{p}-\boldsymbol{p}_{ref}}\right|\right|}
\end{equation}

% The third one is the centroid height guidance reward, which encourages the centroid height $\boldsymbol{h}$ of the robot to align with the centroid height $\boldsymbol{h}_{ref}$ of the reference trajectory.

% \begin{equation}
%     r_{height} = \left|\left|{\boldsymbol{h}-\boldsymbol{h}_{ref}}\right|\right|^2_2
% \end{equation}

\noindent\textbf{Task Reward \& Regularization Reward.}
The task reward encourages the humanoid robot to robustly track the user velocity command, including both the linear velocity and angular velocity.
The regularization reward regularizes the RL policy.
Let $\boldsymbol{v}_{xy}$ be the base linear velocity, $\boldsymbol{c}_{xy}$ be the command linear velocity, $\boldsymbol{w}$ be the base angular velocity, $\boldsymbol{c}_{w}$ be the command angular velocity, $\boldsymbol{\tau}$ be the motor torque, $\boldsymbol{q}_{default}$ be the default joint position, $T_{air}$ be the feet air time, $\mathds{1}$(foot contact) be foot contact with the ground, $\mathds{1}$(collision) be self collision, and $\mathds{1}$(termination) be termination.
These reward components are detailed in Tab.~\ref{tab:reward}.
% along with their respective formulations and weights.
%  for energy efficiency
% The task is to track the root velocity of user command. Therefore, the task reward is composed of root velocity tracking reward, base orientation reward, action rate reward, collision reward, default joint pos

\begin{table}
\centering
\small
\vspace{6pt}
\scalebox{0.85}{
\begin{tabular}{@{}lcccc@{}}
\toprule
Task Reward & Definition & Weights\\
\midrule
% \hline
Linear Velocity & $e^{-4·(\boldsymbol{v}_{xy}-\boldsymbol{c}_{xy})^2}$ & 3.0\\
Angular Velocity & $e^{-4·(\boldsymbol{w}-\boldsymbol{c}_{w})^2}$ & 2.5\\
\midrule
% \toprule
\midrule
Regularization Reward & Definition & Weights\\
\midrule
Z Axis Linear Velocity & $\sum \boldsymbol{v}_z^2$ & -0.8\\
X, Y Axis Angular Velocity &  $\sum \boldsymbol{w}_{xy}^2$ & -0.05\\
Projected Gravity & $\sum \boldsymbol{g}_{xy}^2$ & -6.0\\
Torque  & $\sum\boldsymbol{\tau}^2$ & -5e-6\\
DoF Velocity   & $\sum\dot{\boldsymbol{q}}^2$ & -5e-4\\
DoF Acceleration  & $\sum\ddot{\boldsymbol{q}}^2$ & -2e-8\\
Action Rate & $\sum\dot{\boldsymbol{a}}^2$ & -0.01\\
Smoothness & $\sum\ddot{\boldsymbol{a}}^2$ & -5e-3\\
% Default Joint Position &  & 0.8\\
Joint Regularization & $\sum(\boldsymbol{q}-\boldsymbol{q}_{default})^2$ & -0.1\\
% Low Speed &  & 0.2\\
% Track Velocity Hard &  & 0.5\\
% Stand Still &  & -3.0\\
Feet Air Time & $T_{air} - 0.5$ & 20\\
% Unbalanced Feet Air Time & & -0.5\\
No Fly & $\mathds{1}$(foot contact) & 0.8\\
Collision & $\mathds{1}$(collision) & -1.0\\
Termination & $\mathds{1}$(termination)  & -200.0\\
\bottomrule
\end{tabular}
}
\small
\caption{Task rewards and regularization rewards.}
\label{tab:reward}
\vspace{-15pt}
\end{table}

\section{EXPERIMENTS}

\subsection{Experimental Setup}

We conduct experiments on NAVIAI, a full-size humanoid robot equipped with 41 degrees of freedom (DoFs). 
We utilize 21 DoFs in the robot's kinematic structure, consisting of 6 DoFs in each leg, 4 DoFs in each arm, and 1 DoF in the waist, while the remaining DoFs are fixed.
The robot stands at 1.65 meters in height with a total mass of 60 kilograms.
To enhance the generalization capability and facilitate successful sim-to-real transfer, we employ domain randomization \cite{tobin2017domain} during training, which involves randomizing key physical parameters within the simulation environment.

\begin{table}
\centering
\vspace{6pt}
\setlength{\tabcolsep}{4pt} % adjust column spacing, default is 6pt
\begin{tabular}{@{}lcccccc@{}}
\toprule
Method & JFID$\downarrow$ & KFID$\downarrow$ & JDTW$\downarrow$ & KDTW$\downarrow$ & MELV$\uparrow$\\
\midrule
% SaW \cite{van2024revisiting} & 3.278 & 0.106 & \\
% PBRS \cite{jeon2023benchmarking} & 2.008 & 0.079 & \\
% PBRS+AMP \cite{peng2021amp} & 1.064 & 0.064 & \\
% HumanMimic \cite{tang2024humanmimic} &  &  & \\
% Ours &\textbf{0.186}& \textbf{0.011} & \textbf{} \\
% SaW \cite{van2024revisiting} & 4.569 & 0.030 & 1770.319 & 167.748 & 2.503 \\
SaW \cite{van2024revisiting} & 4.093 & 0.088 & 1531.796 & 261.786 & 2.503 \\
PBRS \cite{jeon2023benchmarking} & 3.118 & 0.058 & 1554.602 & 210.049 & \textbf{2.590} \\
HumanMimic \cite{tang2024humanmimic} & 2.699 & 0.041 & 1617.444 & 175.792 & 2.363 \\
SaW \cite{van2024revisiting}+AMP \cite{peng2021amp} & 2.712 & 0.065 & 1674.894 & 215.590 & 2.308 \\
PBRS \cite{jeon2023benchmarking}+AMP \cite{peng2021amp} & 2.088 & 0.021 & 1512.808 & 145.697 & 2.297 \\
% HumanMimic \cite{tang2024humanmimic} &  &  \\
% Ours &\textbf{1.024}& \textbf{0.010} \\
Ours & \textbf{0.931} & \textbf{0.004} & \textbf{969.369} & \textbf{67.741} & 2.449\\
\bottomrule
\end{tabular}
\caption{Quantatitive Results.}
\label{tab:quantatitive}
\vspace{5pt}
% \end{table}
%  &  Velocity Accuracy$\downarrow$ 
%  &  Velocity Accuracy$\downarrow$ 
% \begin{table}
\centering
\setlength{\tabcolsep}{5pt} % adjust column spacing, default is 6pt
\begin{tabular}{@{}lcccccc@{}}
\toprule
Method & JFID$\downarrow$ & KFID$\downarrow$ & JDTW$\downarrow$ & KDTW$\downarrow$ & MELV$\uparrow$\\
\midrule
Ours w/o GMP & 2.744 & 0.052 & 1363.294 & 216.294 & 2.479\\
% Ours w/o CMD & 1.014 & 0.015 &\\
Our w/o $r_{dof}$ & 1.338 & 0.012 & 1083.462 & 114.273 & 2.502\\
Our w/o $r_{keypos}$ & 1.065 & 0.015 & 1035.853 & 113.217 & \textbf{2.579}\\
Ours & \textbf{0.931} & \textbf{0.004} & \textbf{969.369} & \textbf{67.741} & 2.449\\
% Our w/o $r_{height}$ & \textbf{0.931} & \textbf{0.008} & \\
% Ours &\textbf{1.024}& \textbf{0.010} & \textbf{} \\
\bottomrule
\end{tabular}
\caption{Ablation study results.}
\label{tab:ablation}
\vspace{-18pt}
\end{table}

\noindent\textbf{Dataset.} Human motion data are collected using the XSenS motion capture system, comprising 37 motion sequences and 47,700 frames recorded at 50 frames per second (fps). To mitigate unbalanced data distribution, the motion data are augmented by mirroring along the X-axis. The data are preprocessed to align with a standard coordinate system, where the z-axis points upward, the x-axis faces forward, and the y-axis corresponds to the left-hand direction. The dataset encompasses diverse human walking motions, including acceleration, deceleration, constant speed, turning, and standing, across various velocities.
% These data are then transformed into the robot's local coordinate system. 

\noindent\textbf{Metric.} To evaluate the naturalness of the humanoid robot's motion, we employ the Fréchet Inception Distance (FID) \cite{heusel2017gans}, a widely adopted metric in motion generation tasks \cite{guo2020action2motion,guo2022generating}. The FID measures the distributional difference between the robot's motions and the retargeted motions derived from human data. The motion features used for FID computation are joint angles and keypoint positions, denoted as JFID and KFID, respectively. By comparing the distribution of the robot's walking motions with the distribution of retargeted motions, we assess the anthropomorphism of the robot's actions. We also compare the Dynamic Time Warping (DTW) \cite{huang2025think} between the robot motions and the expert motions in terms of joint angles (JDTW) and keypoint positions (KDTW). A smaller FID and DTW score indicate greater similarity to human motion. Additionally, we evaluate the velocity tracking performance with Mean Episode Linear Velocity Tracking Reward
(MELV) \cite{cheng2024expressive}.
% , while a larger FID suggests less human-like behavior
% We collect robot motion data at different speeds (0, 0.5 m/s, 1.0 m/s, and 1.5 m/s), each lasting 12 seconds. 
% and motion smoothness with Mean Episode Smoothness Reward (SMO).

\noindent\textbf{Implementation Details.} The hyperparameters $\alpha$, $\beta$, and $\gamma$ for whole-body motion retargeting are set to 1, 1000, and 100, respectively. The user velocity command ranges from 0 to 1.5 m/s in the x-direction, -0.3 to 0.3 m/s in the y-direction, and -0.3 to 0.3 rad/s in the yaw direction. The motion encoder and decoder are implemented as multi-layer perceptrons (MLPs) with hidden sizes of [256, 256] and a latent space dimension of 32. The conditional variational autoencoder is trained with an initial learning rate of 1e-5, which gradually decays to 1e-7 over 240 epochs. The weights for the motion reconstruction loss and KL-divergence loss are set to 1.0 and 1.0, respectively. The command encoder is also implemented as an MLP with hidden sizes of [256, 256, 256]. The actor and critic networks in the RL control policy are MLPs with hidden sizes of [512, 256, 128] and employ the ELU activation function. Training is conducted on a single NVIDIA 4090 GPU (24GB) and takes about 12 hours. The RL control policy is trained using PPO with a learning rate of 1e-3 in Isaac Gym physical simulator \cite{makoviychuk2021isaac}.

\begin{figure}[t]
\centering
\vspace{6pt}
\includegraphics[width=0.9\linewidth]{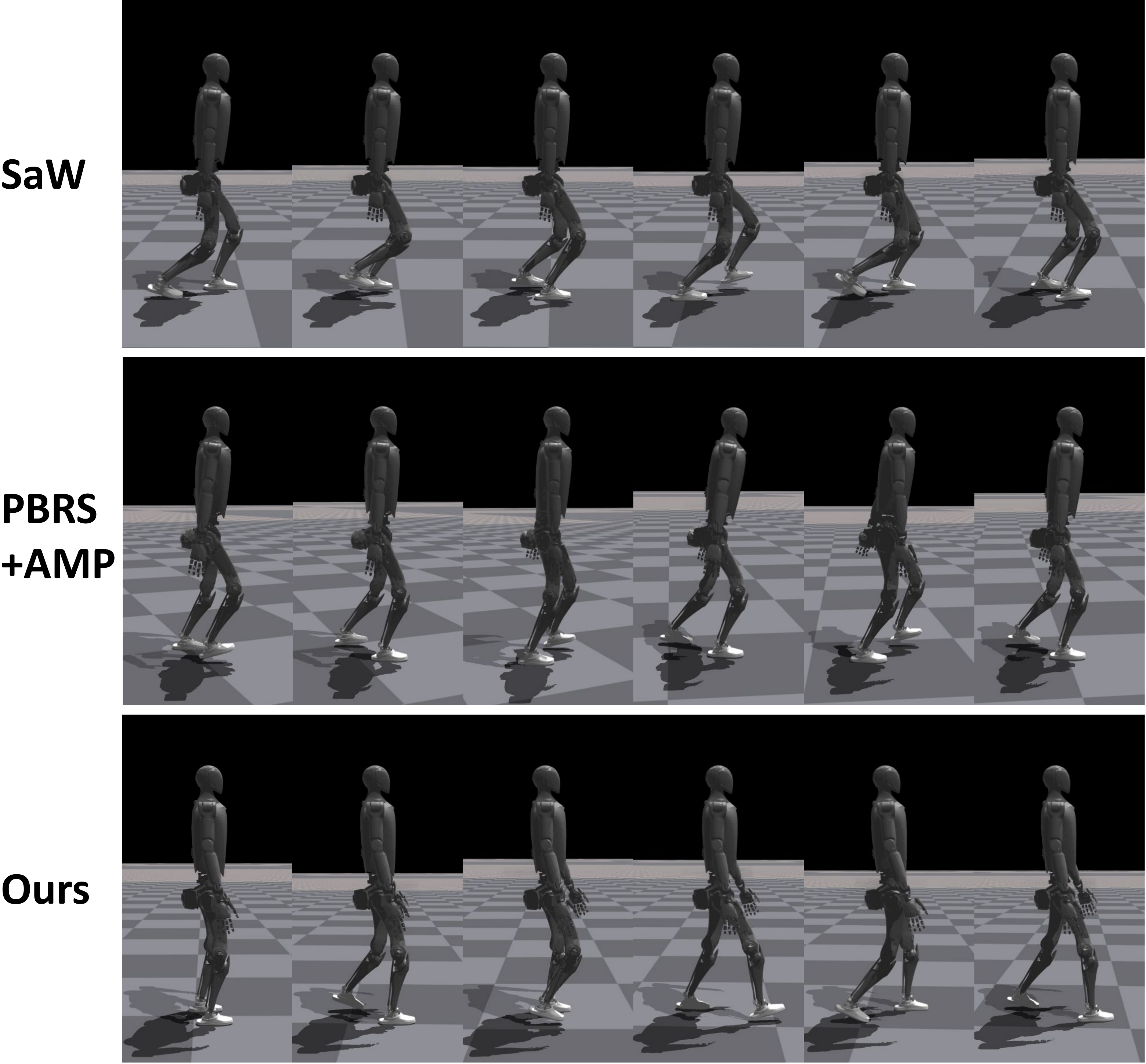}
\caption{Qualitative comparison with representative baselines. The pure RL method (SaW \cite{van2024revisiting}) overlooks motion naturalness, resulting in unnatural bent leg postures. The adversarial method (PBRS \cite{jeon2023benchmarking} + AMP \cite{peng2021amp}) learns straighter legs; however, it relies on ambiguous style rewards and is insufficient to fully capture human-like motion characteristics.
In contrast, our method exhibits superior motion naturalness.
% From the first row to the last row are SaW \cite{van2024revisiting}, PBRS \cite{jeon2023benchmarking} + AMP \cite{peng2021amp}, our method, respectively.
% Snapshots of natural and human-like humanoid robot locomotion in the Isaac Gym simulation environment.
}
\label{fig:simulation}
\vspace{-15pt}
\end{figure}

\begin{figure*}[t]
\centering
\vspace{6pt}
% \includegraphics[width=0.85\linewidth]{images/simulation-experiment.png}
% \caption{Snapshots of natural and human-like humanoid robot locomotion in the Isaac Gym simulation environment.}
% \label{fig:simulation}
% \vspace{-6pt}
% from standing to walking

\includegraphics[width=0.8\linewidth]{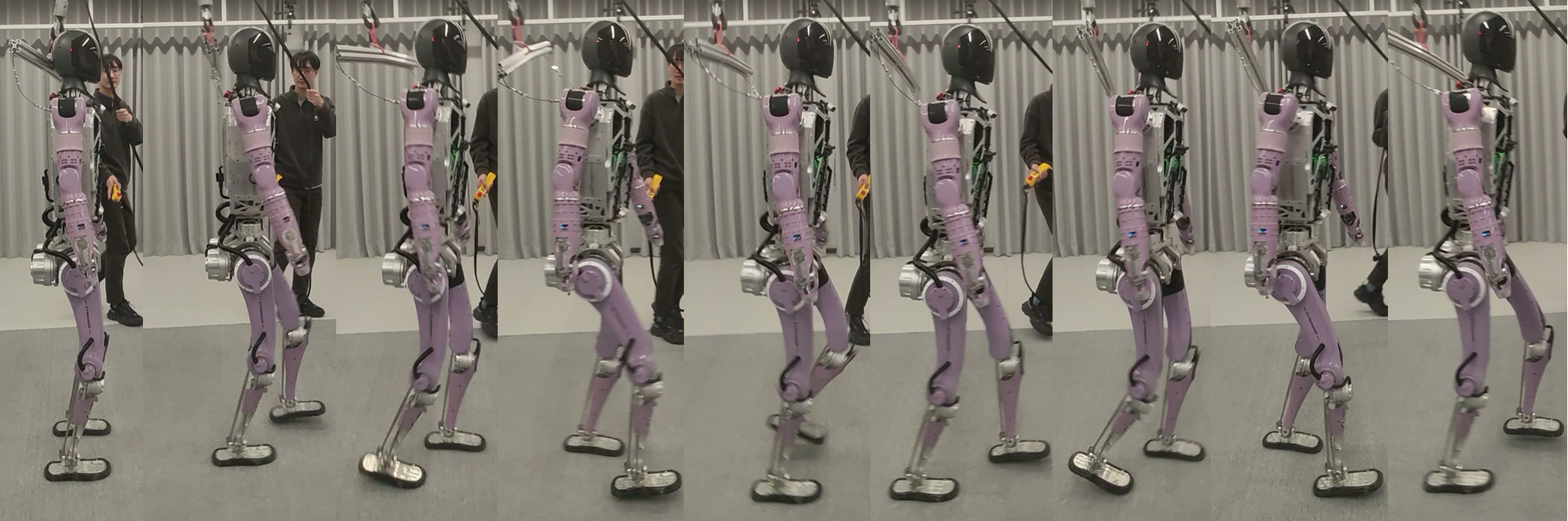}
\caption{Snapshots of natural and human-like humanoid robot locomotion in the real-world environment.}
\label{fig:real-experiment}
\vspace{-18pt}
\end{figure*}
% from standing to walking
% \subsection{User Study}

\subsection{Comparison Study}

To thoroughly evaluate the effectiveness of our method in the task of human-like locomotion, we conduct a comparative analysis against five representative baselines.

\begin{itemize}
    \item SaW \cite{van2024revisiting} is an RL-based standing and walking (SaW) controller using long short-term memory (LSTM) \cite{hochreiter1997long} recurrent neural network.
    % with minimally constraining reward functions to train.
    \item PBRS \cite{jeon2023benchmarking} is an RL-based locomotion policy that utilizes potential based reward shaping to enhance the learning process and achieve faster convergence.
    % employs the Proximal Policy Optimization (PPO) reinforcement learning algorithm to train the bipedal robot for walking. Task-specific rewards are designed, including velocity tracking reward and orientation reward, to guide the learning process.
    \item HumanMimic \cite{tang2024humanmimic} is an RL-based policy that integrates Wasserstein distance and soft boundary constraints to improve stability during adversarial training.
    \item SaW \cite{van2024revisiting} + AMP \cite{peng2021amp} combines SaW with the adversarial motion prior.
    \item PBRS \cite{jeon2023benchmarking} + AMP \cite{peng2021amp} combines PBRS with the adversarial motion prior.
    % reinforcement learning with adversarial neural networks by training a discriminator network to assess the similarity of motion styles, thereby providing style-based rewards for reinforcement learning.
\end{itemize}

The quantitative results presented in Tab.~\ref{tab:quantatitive} demonstrate that our method significantly outperforms baseline approaches in motion naturalness. Pure RL-based methods, including SaW and PBRS, exhibit the lowest motion naturalness despite achieving relatively higher velocity tracking accuracy, as they primarily focus on velocity tracking while neglecting the naturalness of motion. Adversarial-based methods, including SaW+AMP, PBRS+AMP, and HumanMimic, show improved performance compared to pure RL methods by leveraging human reference datasets to learn human-like motion styles. However, these methods are constrained by the inherent instability of adversarial training and the risk of mode collapse, often resulting in the learning of motion styles specific to particular reference data. In contrast, our proposed method employs a generative motion prior to dynamically generate robot reference motions in response to velocity commands, thereby providing sufficient and adaptive guidance for the RL control policy learning. This approach ensures both task performance and motion naturalness, addressing the limitations of existing methods.

\subsection{Ablation Study}

\noindent\textbf{Generative Motion Prior.} To evaluate the effectiveness of the generative motion prior in motion guidance, we compare our full method with a variant that excludes this component, denoted as ``Ours w/o GMP''. 
As shown in Tab.~\ref{tab:ablation}, the absence of the generative motion prior leads to a significant decline in motion naturalness. This underscores the critical role of the generative model in providing fine-grained guidance during the learning process.

% The results in Tab.~\ref{tab:ablation} demonstrates that the motion naturalness decreases significantly without the generative motion prior, indicating the importance of leveraging a generative model to guide the learning process.

\noindent\textbf{Motion Guidance Reward.} To further investigate the contribution of individual motion guidance rewards, evaluate variants of our method that exclude the joint angle guidance reward and the keypoint position guidance reward, denoted as ``Ours w/o $r_{dof}$'' and ``Ours w/o $r_{keypos}$'', respectively. The results in Tab.~\ref{tab:ablation} demonstrate that the removal of either guidance reward results in a noticeable performance degradation. This highlights the importance of granular motion guidance at both the joint angle and keypoint position levels in preserving the naturalness and human-like quality of the generated motions.

% \noindent\textbf{Command Encoder.}

% \noindent\textbf{}

\subsection{Qualitative Study}

\begin{figure}[t]
\centering
% \vspace{6pt}
\includegraphics[width=0.7\linewidth]{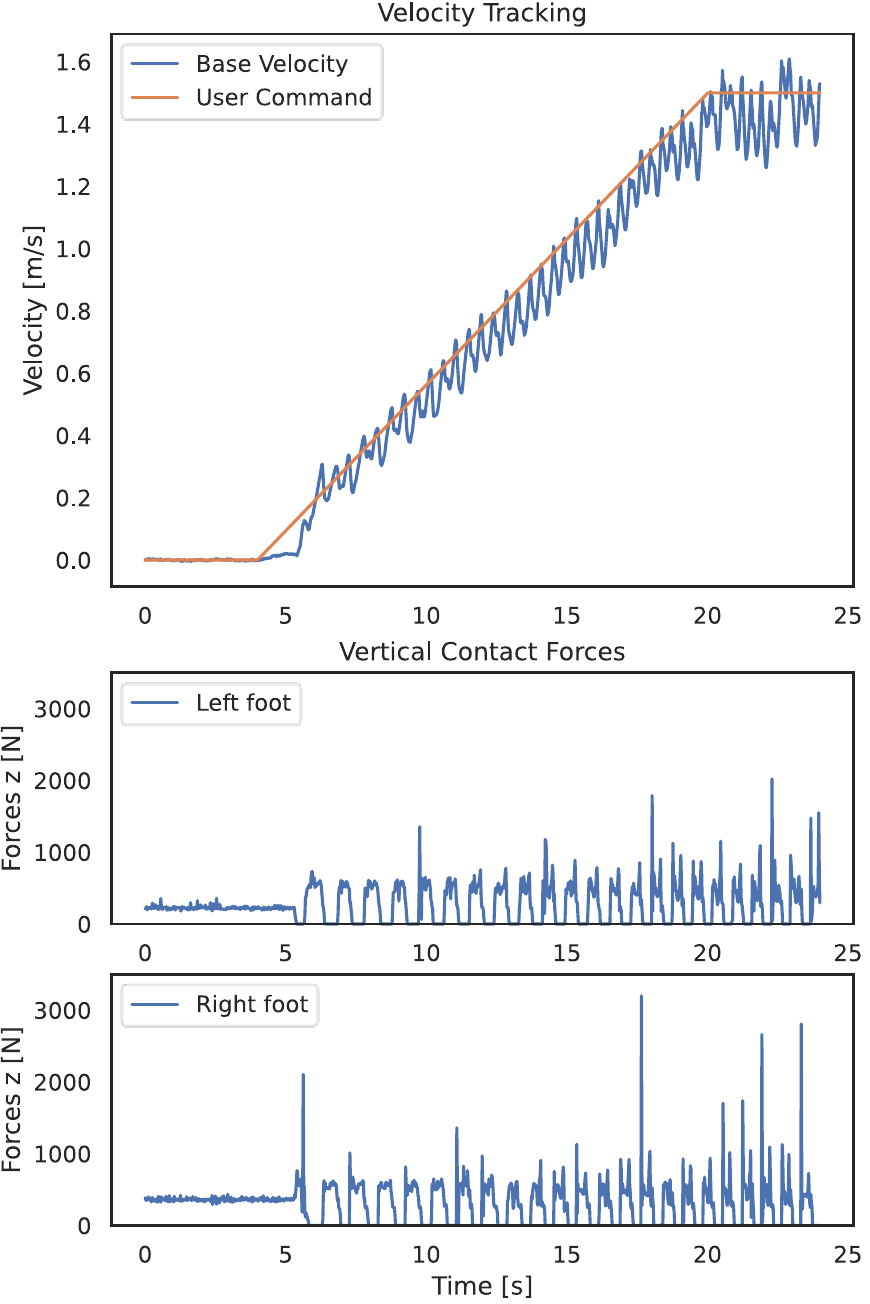}
\caption{Velocity tracking results and vertical foot contact forces from standing to walking in 1.5m/s.}
\label{fig:velocity-tracking}
\vspace{-15pt}
\end{figure}

\noindent\textbf{Velocity Tracking.}
As illustrated in Fig.~\ref{fig:velocity-tracking}, we present the velocity tracking performance alongside the corresponding vertical foot contact forces during the transition from standing to walking at a target velocity of 1.5 m/s. The experimental results demonstrate that our proposed method achieves robust and precise tracking of the user-specified velocity commands while ensuring a smooth and natural transition between standing and walking states.
Fig.~\ref{fig:simulation} demonstrates that the learned controller performs well in the Isaac Gym simulation environment and the humanoid robot walks in a more natural and human-like motion pattern compared with baseline methods, which highlights the effectiveness of our approach in generating human-like locomotion patterns.
% stable
% This capability is further validated by the consistent and physically plausible foot contact forces observed throughout the motion, highlighting the effectiveness of our approach in generating stable and human-like locomotion patterns.

% \noindent\textbf{Sim-to-Sim Experiment.}
% To validate the robustness of the proposed method, we conduct a sim-to-sim experiment which deploys the policy trained in Isaac simulator to Gazebo simulator. Fig.~\ref{} demonstrates that the learned controller performs well in the Gazebo simulation environment and the humanoid robot walks in a natural and human-like motion pattern, suggesting that the proposed method is robust in different physical simulation.

\noindent\textbf{Real-world Experiment.}
To verify the effectiveness of our proposed method, we conduct comprehensive real-world experiments utilizing the NAVIAI full-sized humanoid robot platform. As illustrated in Fig.~\ref{fig:real-experiment}, the humanoid robot demonstrates robust tracking performance in response to user velocity commands while maintaining a high degree of motion naturalness that closely resembles human movement characteristics.

% These experimental results provide compelling evidence for the effectiveness and practical applicability of our proposed approach in real-world humanoid robot control scenarios.

% \begin{table}[]
% \begin{tabular}{llllllllllll}
% \hline
% \multirow{2}{*}{Method} & \multicolumn{3}{c}{FID$\downarrow$}              & \multicolumn{4}{c}{Joint Angle FID$\downarrow$}          & \multicolumn{4}{c}{Keypoint FID$\downarrow$}            \\ \cline{2-12} 
%                         & 0.5m/s         & 1.0m/s         & 1.5m/s         & \multicolumn{2}{l}{0.5m/s} & 1.0m/s       & 1.5m/s       & \multicolumn{2}{l}{0.5m/s} & 1.0m/s       & 1.5m/s      \\ \hline
% RL                      & 3.092          & 3.460          & 3.729          & \multicolumn{2}{l}{}       &              &              & \multicolumn{2}{l}{}       &              &             \\
% RL+AMP                  &                &                &                & \multicolumn{2}{l}{}       &              &              & \multicolumn{2}{l}{}       &              &             \\
% RL+W-AMP                & 1.864          & 2.067          & 2.241          & \multicolumn{2}{l}{}       &              &              & \multicolumn{2}{l}{}       &              &             \\
% Ours                    & \textbf{0.175} & \textbf{0.190} & \textbf{0.255} & \multicolumn{2}{l}{}       &              &              & \multicolumn{2}{l}{}       &              &             \\ \hline
% \end{tabular}
% \caption{Quantatitive Results.}
% \end{table}

\section{CONCLUSIONS}

In this paper, we introduce a generative motion prior framework designed to enable natural, human-like locomotion for humanoid robots. Our approach leverages a generative model to predict future robot reference motion trajectories, providing granular and interpretable guidance for the RL-based control policy. The generative motion prior serves as a frozen expert model, ensuring stable and high-quality motion trajectory-level guidance throughout the training process. Extensive experiments conducted in both simulated and real-world environments demonstrate the effectiveness of our method in generating natural and plausible locomotion for humanoid robots. This work highlights the potential of combining generative models with RL to achieve human-like motion generation in robotics.
 % robust and
% \vspace{-5pt}
\printbibliography

\addtolength{\textheight}{-12cm}   % This command serves to balance the column lengths
                                  % on the last page of the document manually. It shortens
                                  % the textheight of the last page by a suitable amount.
                                  % This command does not take effect until the next page
                                  % so it should come on the page before the last. Make
                                  % sure that you do not shorten the textheight too much.

%%%%%%%%%%%%%%%%%%%%%%%%%%%%%%%%%%%%%%%%%%%%%%%%%%%%%%%%%%%%%%%%%%%%%%%%%%%%%%%%

%%%%%%%%%%%%%%%%%%%%%%%%%%%%%%%%%%%%%%%%%%%%%%%%%%%%%%%%%%%%%%%%%%%%%%%%%%%%%%%%

%%%%%%%%%%%%%%%%%%%%%%%%%%%%%%%%%%%%%%%%%%%%%%%%%%%%%%%%%%%%%%%%%%%%%%%%%%%%%%%%
% \section*{APPENDIX}

% Appendixes should appear before the acknowledgment.

% \section*{ACKNOWLEDGMENT}

% The preferred spelling of the word ÒacknowledgmentÓ in America is without an ÒeÓ after the ÒgÓ. Avoid the stilted expression, ÒOne of us (R. B. G.) thanks . . .Ó  Instead, try ÒR. B. G. thanksÓ. Put sponsor acknowledgments in the unnumbered footnote on the first page.

%%%%%%%%%%%%%%%%%%%%%%%%%%%%%%%%%%%%%%%%%%%%%%%%%%%%%%%%%%%%%%%%%%%%%%%%%%%%%%%%

% \begin{thebibliography}{99}

% \end{thebibliography}

\end{document}